\documentclass[conference]{IEEEtran}
\IEEEoverridecommandlockouts

\usepackage{cite}
\def\BibTeX{{\rm B\kern-.05em{\sc i\kern-.025em b}\kern-.08em
    T\kern-.1667em\lower.7ex\hbox{E}\kern-.125emX}}
\usepackage{amsthm}
\newtheorem{definition}{Definition}[section]
\newtheorem{proposition}{Proposition}[section]

\usepackage{amsfonts}
\usepackage{amsmath}
\usepackage{bm}
\usepackage{subcaption}

\usepackage{tikz}
\usetikzlibrary{calc,shapes,positioning}
\pgfdeclarelayer{background}
\pgfsetlayers{background,main}

\usepackage{textgreek}

\usepackage[linesnumbered,ruled]{algorithm2e}
\usepackage{multirow}

\usepackage{comment}
\usepackage{url}

\def\X{$\times$}
\newcommand{\citep}[1]{\cite{#1}}
\newcommand{\toprule}{\hline\hline}
\newcommand{\midrule}{\hline}
\newcommand{\bottomrule}{\hline\hline}

\def\makeheadbox{\relax}

\title{
    \textDelta-Closure Structure for Studying Data Distribution
    \thanks{Identify applicable funding agency here. If none, delete this.}
}

\author{
    \IEEEauthorblockN{Aleksey Buzmakov}
    \IEEEauthorblockA{
        \textit{National Research University Higher School of Economics}\\
        Perm, Russia \\
        avbuzmakov@hse.ru
    }
\vspace{5mm}    
    \IEEEauthorblockN{Tatiana Makhalova}
    \IEEEauthorblockA{
        \textit{Universit\'e de Lorraine, CNRS, Inria, LORIA}\\
        54000 Nancy, France \\
        t.makhalova@gmail.com
    }

\and
    \IEEEauthorblockN{Sergei O. Kuznetsov}
    \IEEEauthorblockA{
        \textit{National Research University Higher School of Economics}\\
        Moscow, Russia \\
        skuznetsov@hse.ru
    }
\vspace{5mm}    
    \IEEEauthorblockN{Amedeo~Napoli}
    \IEEEauthorblockA{
        \textit{Universit\'e de Lorraine, CNRS, Inria, LORIA}\\
        54000 Nancy, France \\
        amedeo.napoli@loria.fr
    }
}

\begin{document}

\maketitle


\begin{abstract}
%
In this paper, we revisit pattern mining and study the distribution underlying a binary dataset thanks to the closure structure which is based on passkeys, i.e., minimum generators in equivalence classes robust to noise.
We introduce $\Delta$-closedness, a generalization of the closure operator, where $\Delta$ measures how a closed set differs from its upper neighbors in the partial order induced by closure.
A $\Delta$-class of equivalence includes minimum and maximum elements and allows us to characterize the distribution underlying the data.
Moreover, the set of $\Delta$-classes of equivalence can be partitioned into the so-called $\Delta$-closure structure.
In particular, a $\Delta$-class of equivalence with a high level demonstrates correlations among many attributes, which are supported by more observations when $\Delta$ is large.
In the experiments, we study the $\Delta$-closure structure of several real-world datasets and show that this structure is very stable for large $\Delta$ and does not substantially depend on the data sampling used for the analysis.

\end{abstract}

\section{Introduction}\label{sec:introduction}
%

%
%
%
%
%
%
%

In this paper, we are interested in pattern or itemset mining in tabular data.
There is a considerable work on this subject, especially regarding algorithms and search for interesting patterns \cite{AggarwalH14}.
Here we rather focus on the distribution underlying the dataset under study thanks to closed itemsets, their equivalence classes, and the associated generators.

Many pattern mining approaches produce  a particular set of paterns, which provides a ``certain view'' of the intrinsic structure underlying the data. However, this view is not usually directly related to the distribution underlying the data.
In this paper we propose a methodology for computing and understanding the intrinsic structure of a dataset, based on the recently introduced \emph{closure structure}~\cite{MakhalovaBKN22}.
The closure structure reveals the distribution of the itemsets in the data in terms of frequency, and also of stability (how a closed set depends on its content), and in addition supports an interpretation of the content of a dataset.
In this paper we propose a methodology for computing and understanding the intrinsic structure of a dataset in two main ways:
(i) we introduce the \emph{closure structure}, which reveals the distribution of the itemsets in the data in terms of frequency,
(ii) we generalize the closure structure to \textDelta-\textit{closure structure}, which is a more robust closure structure related to the distribution underlying the data.
These two closure structures support an interpretation of the content of a dataset.
The closure structure is based on ``closed itemsets'' and minimum elements of their ``equivalence classes'', which are computed independently of any interestingness measure or set of constraints.
The closure structure and its levels provide a representation of the complexity of the content of a dataset.
We propose a formalization of the closure structure in terms of Formal Concept Analysis \citep{GanterW99}, which is well adapted to the study of closed itemsets, equivalence classes, and data topology in terms of closed sets.
We also study how sampling and the closure structure are related, for possibly dealing with large datasets.

Then we generalize the notion of closure to $\Delta$-closedness which allows us to work with stability along with frequency.
Actually, $\Delta$ measures how much a closed set differs from its upper neighbors in the partial order of closed sets.
A $\Delta$-class of equivalence allows us to characterize the distribution underlying the data:
(i) when $\Delta$ is large, there are only a few $\Delta$-classes of equivalence whose elements are very stable,
(ii) when $\Delta$ is small, the number of $\Delta$-classes increases while the related information is less stable and depends on a smaller number of elements.
This allows us to study stable patterns which are robust against noise and also to dynamic changes in the data.

In the experiments, we compute the closure structure of a number of public datasets and show how $\Delta$-closedness is taken into account.
The closure structure may roughly take three types, where the first levels of the structure are the most interesting, stable, and interpretable in standard and almost standard datasets, while in non-standard datasets, levels are more separated and less easily interpretable.
For example, the closure structure allows us to determine the levels where the itemsets are most diverse.
It shows how the frequency and the stability are distributed among the levels, and when a search for interesting itemsets can be stopped without major loss.
Moreover, the $\Delta$-closure structure is very stable for large $\Delta$ and does not substantially depend on the data sampling used for the analysis.
To the best of our knowledge, such a study and a visualization of the content of a dataset in terms of closed itemsets is rather unique and may guide the mining of a dataset and the interpretation of the resulting itemsets.

The paper has the following structure.
Section~\ref{sec:related_work} presents related work and motivation, while in Section~\ref{sec:basics} we recall some basic notions and we also introduce the closure levels, equivalence classes, and passkeys for the standard closure operation.
In Section~\ref{sec:closure+sampling} we study the behavior of the closure structure under sampling and the properties of closure and passkeys.
In Section~\ref{sec:delta-operator} we introduce the $\Delta$-closedness and the related equivalence classes and passkeys.
In Section~\ref{sec:experiments} we discuss the closure structure and the stability of passkeys, and we show how the closure structure can be visualized and interpreted.
Then we conclude the paper and give directions of future work.

%
\section{Related work and Motivation}
\label{sec:related_work}
%

Following the research directions discussed in \cite{BoleyHW09,MakhalovaBKN22}, we are interested in this paper in discovering closed and stable patterns together with their related equivalence classes and minimum generators (passkeys).
As in \cite{BoleyHW09}, we introduce a $\Delta$ parameter for studying $\Delta$-closure and the robustness of the discovered patterns against noise, and as well the stability of the associated equivalence classes and their characteristics elements such as passkeys.
Moreover, as in \cite{MakhalovaBKN22} we are also interested in investigating the ``data distribution'' through the so-closed ``closure structure''.
The objective is to build a kind of ``internal picture'' of the datasets under study to check at which level --related to closure and $\Delta$-closure-- one can discover the most interesting patterns, and as well interesting implications or association rules.
In addition, in the present paper, we introduce \textDelta-closure structure based  on \textDelta-closure operator which subsumes ordinary closure.

The interest in data distribution is frequently appearing in the literature.
Indeed, even the standard frequent itemset mining indirectly focuses on properties of the distribution.
Such distribution interest can be stated more formally if the itemset frequency is verified as a statistical hypothesis~\cite{RiondatoV14}.
Some works study how to take noise into account in pattern mining, which can also be considered as indirect characterization of the distribution underlying the data.
In particular, \textdelta-free sets~\cite{BoulicautBR03} are special classes of patterns that encodes association rules with high confidence (only few counter-examples are allowed) allowing identification of noise-resistant patterns~\cite{PensaB05}.
Meanwhile other studies were performed to find out classes of patterns resistant to noise~\cite{YangFB01,KlimushkinOR10}.

Still about data distribution, in \cite{RameshMZ03} the authors study distribution of sizes of frequent and maximal frequent itemsets in a database.
In \cite{FlouvatMP10} the authors are mostly interested in the distributions related to the border between frequent and infrequent itemsets together with the distribution of three concise representations:
frequent closed, frequent free and frequent essential itemsets.
In \cite{LeeuwenU14} the authors propose an efficient approach to estimation of the number of frequent patterns for arbitrary minsup thresholds.
In our study we assume that ``important’’ patterns enabling better interpretation are not necessarily frequent, but are certainly stable to noise in data.
So, here we study empirical distribution of noise tolerance in terms of $\Delta$-stability introduced below.

Another challenge in itemset mining remains computational complexity.
The number of itemsets can be exponential in the dataset size.
Focusing on closed itemsets allows for a substantial reduction of this number by replacing an equivalence class, i.e., a whole class of itemsets having the same support, by the largest one which is the closed itemset~\citep{PasquierBTL99}.
For dealing with large amounts of closed itemsets, it is possible to evaluate itemsets thanks to quality metrics~\cite{VreekenT14}.
For example $\Delta$-closedness evaluates the robustness of patterns and the corresponding closure operation~\cite{BoleyHW09}.
Roughly speaking, a $\Delta$-closed set cannot be augmented by any further item without decreasing its support by at least $\Delta$.
Actually, $\Delta$-closed sets allow to capture interesting itemsets that are stable, i.e., robust to noise and changes in the data.
A very similar concept is \textdelta-tolerance closed itemsets which are defined w.r.t. itemset support rather than an absolute change in support for \textDelta-closure~\cite{ChengKN06}.
An alternative to exhaustive enumeration of itemsets is based on ``sampling'' \citep{DzyubaLR17} and on a gradual search for itemsets according to an interestingness measure or a set of constraints \citep{SmetsV12}.
Such algorithms usually output a rather small set of itemsets while they may provide only an approximate solution.


The papers discussed above do not address the problem of discovering an intrinsic structure underlying the set of closed itemsets, and -- more generally -- the dataset itself.
By contrast, in this paper, we define such a level-wise structure for representing this intrinsic structure of the datasets.
Moreover, we build this structure by means of $\Delta$-closedness operator and we study the stability of the related $\Delta$-closure levels w.r.t. the size of $\Delta$.
The \textDelta-closure reveals the internal structure of a dataset and as well relates this structure to the distribution underlying the dataset.

\section{Formalism} \label{sec:basics}
%


A dataset as modeled as a \textit{formal context}~\citep{GanterW99}, which is a triple $\mathbb{K}=\left ( G,M,I \right )$, where $G$ is a set of objects, $M$ is a set of attributes and $I\subseteq G\times M$ is an incidence relation such that $\left ( g, m \right )\in I $ if object $g$ has attribute $m$. 
Two derivation operators $\left ( \cdot \right )' $ are defined for $A\subseteq G$ and $B \subseteq M$ as follows:
\begin{align}
A' &= \left \{ m \in M \mid \forall  g \in A : gIm  \right \} \\
B' &= \left \{ g \in G \mid  \forall  m \in B : gIm  \right \}
\end{align}

Intuitively, $A'$ is the set of common attributes to objects in $A$, while $B'$ is the set of objects which have all attributes in $B$.
Sets $A \subseteq G$, $B \subseteq M$, such that $A = A''$ and $B = B''$, are closed sets and $(\cdot)''$ is a closure operator being equivalent to instance counting closure in data mining~\cite{PasquierBTL99}.
For $A \subseteq G$, $B \subseteq M$, a pair $(A,B)$ such that $A'=B$ and $B'=A$, is called a \textit{formal concept}, then $A$ and $B$ are closed sets and called \textit{extent} and \textit{intent}, respectively. 

%
%


\newcommand{\K}{\mathbb{K}}
\newcommand{\T}{G}
\newcommand{\A}{M}
\newcommand{\R}{I}

\newcommand{\attr}{m}
\newcommand{\obj}{g}
\newcommand{\set}[1]{\{#1\}}

\newcommand{\tikznamedpicture}[3][0]{
	\newcommand{#2}[#1]{ \begin{tikzpicture}#3\end{tikzpicture} }
}

\newcommand{\concept}[2]{$\left(#1;#2\right)$}

\tikznamedpicture{\putStabLattice}{[
	every node/.style={draw,rectangle,font={\tiny}},
	node distance= 0.5cm and 0.3cm
	]
	\node(g5){\concept{5}{abcf}};
	\node(g6)[right=of g5]{\concept{6}{abcg}};
	\node(g7)[right=of g6]{\concept{7}{abdeh}};
	\node(g8)[right=of g7]{\concept{8}{abdei}};
	\node(g12)[above=of $(g6)!0.5!(g7)$] {\concept{12}{abcde}[2]};
	\node(bottom)[below=of $(g6)!0.5!(g7)$] {\concept{\emptyset}{\A}}
	edge(g5)
	edge(g6)
	edge(g7)
	edge(g8)
	edge(g12);
	\node(g123)[above=of g12] {\concept{123}{abcd}}
	edge(g12);
	\node(g124)[right=of g123] {\concept{124}{abce}}
	edge(g12);
	\node(g1278)[right=of g124] {\concept{1278}{abde}[2]}
	edge(g7)
	edge(g8)
	edge(g12);
	\node(g123456)[above left=of g123] {\concept{123456}{abc}[3]}
	edge(g5)
	edge(g6)
	edge(g123)
	edge(g124);
	\node(g12378)[right=of g123456] {\concept{12378}{abd}}
	edge(g123)
	edge(g1278);
	\node(g12478)[right=of g12378] {\concept{12478}{abe}}
	edge(g124)
	edge(g1278);
	\node(g12345678)[above=of g12378] {\concept{12345678}{ab}[2]}
	edge(g123456)
	edge(g12378)
	edge(g12478);
	\node(g123456789)[above left=of g12345678] {\concept{123456789}{a}}
	edge(g12345678);
	\node(g123456780)[above right=of g12345678] {\concept{123456780}{b}}
	edge(g12345678);
	\node(top)[above=of $(g123456789)!0.5!(g123456780)$] {\concept{1234567890}{\emptyset}}
	edge(g123456789)
	edge(g123456780);
}

\begin{figure}[t]
  \centering
    \resizebox{\columnwidth}{!}
    {
    \begin{tabular}{l|ccccccccc}
      \hline\hline
      & $\attr_1$ & $\attr_2$ & $\attr_3$ & $\attr_4$ & $\attr_5$ & $\attr_6$ & $\attr_7$ & $\attr_8$ & $\attr_9$ \\
      & $a$ & $b$ & $c$ & $d$ & $e$ & $f$ & $g$ & $h$ & $i$ \\
      \hline
      $\obj_1$ &\X&\X&\X&\X&\X& & & & \\
      $\obj_2$ &\X&\X&\X&\X&\X& & & & \\
      $\obj_3$ &\X&\X&\X&\X& & & & & \\
      $\obj_4$ &\X&\X&\X& &\X& & & & \\
      $\obj_5$ &\X&\X&\X& & &\X& & & \\
      $\obj_6$ &\X&\X&\X& & & &\X& & \\
      $\obj_7$ &\X&\X& &\X&\X& & &\X& \\
      $\obj_8$ &\X&\X& &\X&\X& & & &\X\\
      $\obj_9$ &\X& & & & & & & & \\
      $\obj_{10}$& &\X& & & & & & & \\
      \hline\hline 
    \end{tabular}
    } 
    
  \caption{
    A toy binary dataset, `x' shows if an object is related to an attribute.
    \label{fig:stability-context}
   }
\end{figure}

    \begin{proposition}\label{prop:sample-passkey}
        If an itemset $X$ is a passkey in a sample dataset $\mathbb{K}_s$ then it is a passkey in the whole dataset $\mathbb{K}$.
    \end{proposition}

    Propositions~\ref{prop:sample-itemset},~\ref{prop:sample-level},~\ref{prop:sample-key} and~\ref{prop:sample-passkey} describe how the closure structure changes in a sample.
    In particular, some closed itemsets can disappear from the closure structure in a sample.
    Some other can change its level.
    Let us consider closure structures for samples in Figure~\ref{fig:sampled_structures}.
    The itemset $abde$ is closed with $G^{(1)}_s$, but is not closed with $G^{(2)}_s$ and, thus, it is absent in the right closure structure.
    The passkey $de$ of $abde$ in $G^{(1)}$ becomes the passkey of $abcde$ in the sample $G^{(2)}$.
    In fact, in $G^{(2)}$ the itemsets $abde$ and $abcde$ belong to the same class of equivalence with the closure being equal to $abcde$.
    It changes the level of $abcde$ from 3 (the passkey is $cde$ in the first sample) to 2 (the passkey is $de$ in the second sample).
    This change can be considered in two ways.
    If every element in the closure structure is associated with the closure, then the itemset $abcde$ ``jumps'' from level 3 to level 2.
    However, if the elements in the closure structure are associated with the corresponding passkeys, then there is no ``jump''.
    Every element of the structure is either preserved or removed in a sample.
    If an element is preserved, its closure can change.
    In particular, in the previous example both closure structures have an element associated with the passkey $de$.
    Nevertheless, the closure of the passkey is different in the two samples above, i.e., $abde$ and $abcde$ in first and second samples, respectively.

    We have seen that under sampling some elements of the closure structure can disappear.
    Nevertheless, there are some elements that are more frequently preserved in a sample than others.
    There are various approaches to measure the probability of an element to be preserved under sampling. In particular, stability~\cite{Kuznetsov07} and robustness~\cite{TattiMC14} measure the probability of a pattern to remain closed under sampling. Both of them are specializations of itemset closedness.
    
    \begin{definition}\label{def:sample-closedness}
      Let $w: 2^{\T} \rightarrow \mathbb{R}_{+}$ be a weighting function of subdatasets.
      Then the \textit{closedness} of an itemset $X$ w.r.t. $w$ is the total relative weight of the subdatasets where the itemset $X$ is closed:\begin{equation}\label{eq:sample-closedness}
       \frac{\sum\{w(S)\mid S \subseteq \T \wedge X \text{ is closed in } (S,\A,\R)\}}{\sum\{w(S)\mid S \subseteq \T\}}.
      \end{equation}
    \end{definition}
    
    \emph{Stability} is a special case of closedness when the weights $w$ of  all subdatasets $\K_s$ of $\K$ are equal, i.e., $w(\K_s)=2^{-|\T|}$.
    In this case we consider every subdataset equally probable and compute the probability that the itemset $X$ is closed.
    \emph{Robustness}~\cite{TattiMC14} is also a special case of itemset closedness.
    Additionally, Definition~\ref{def:sample-closedness} can be modified and applied to other elements of the pattern space including keys and passkeys.
    For that in~(\ref{eq:sample-closedness}) one should replace the verification ``if $X$ is closed in a subdataset'' with ``if $X$ is a key or a passkey''.
    
    Accordingly, when studying a dataset, one can be interested in the more ``stable'' part the closure structure, which is less dependent on the dataset itself and more related to the distribution supporting the dataset.
    As we will see in the next section, passkey-stability as well as closure-stability are related to the \textDelta-closure operator and the corresponding class of equivalence.

\end{comment}

%
%

%
\subsection{\textDelta-classes of equivalence}
%

\begin{definition}
    An itemset $B$ is called \textDelta-closed if for any attribute $m \in M$:
    \begin{equation}
        \label{eq:closedness-delta}
        |B'| - |(B \cup \{m\})'| \geq \Delta \geq 1.
    \end{equation}
\end{definition}

If $\Delta = 1$, then the itemset is just \emph{closed} w.r.t. object counting~\cite{PasquierBTL99,BoleyHW09}.
In~\cite{BoleyHW09} it was shown that \textDelta-closedness is associated with a closure operator (we call it \textDelta-closure).
From a computational point of view \textDelta-closure works as the following. Given a non \textDelta-closed itemset $B$, i.e., 
$\exists m\in M(|B'| - |(B \cup \{m\})'| < \Delta)$, 
it can be closed by iteratively changing $B$ to $B \cup \{m\}$, for any $m$ violating~(\ref{eq:closedness-delta}) until no such attribute is found. The result is the \textDelta-closure of $B$. The corresponding closure operator is denoted by $\phi_\Delta$. Any closure operator is associated with a class of equivalence.

\begin{definition}\label{def:equiv-d}
    Given an itemset $X$, its equivalence class $Equiv_\Delta(X)$ is the set of all itemsets with closure equal to the closure of $X$, i.e., 
    \begin{equation}
        \label{eq:equiv-delta}
        Equiv_\Delta(X) = \{Y \subseteq M \mid \phi_\Delta(Y)=\phi_\Delta(X)\}.
    \end{equation}
\end{definition}

Some elements of a class of equivalence can be highlighted since they have special properties~\cite{MakhalovaBKN22}.

\begin{definition}
A \emph{\textDelta-key} $X \in Equiv_\Delta(B)$ is any minimal (w.r.t. subset relation) itemset in $Equiv_\Delta(B)$. We denote the set of \textDelta-keys (\textit{\textDelta-key set}) of $B$ by $Key_\Delta(B)$. \label{def:key-d}
\end{definition}

\begin{definition}
An itemset $X \in Key_\Delta(B)$ is called a \emph{\textDelta-passkey} if it has the smallest size among all keys in $Key_{\Delta}(B)$. We denote the set of passkeys by $pKey_\Delta(B) \subseteq Key_\Delta(B)$. For a \textDelta-closed itemset $B$ the \emph{\textDelta-passkey set} is given by $pKey_\Delta(B) = \{ X \mid X \in Key_\Delta(B), |X| = min_{Y \in Key_\Delta(B)}|Y|\}$. 
\label{def:min_key-d}
\end{definition}

The notions of \textDelta-closure and \textDelta-(pass)keys are exemplified in Section~\ref{sect:example}.
It can be seen that \textDelta-classes of equivalence for high $\Delta$ are partitions of some \textDelta-classes of equivalence for smaller \textDelta, i.e., 
$
(\forall \Delta\geq 2)(\forall X \subseteq M)(\exists X_1,\dots,X_k)Equiv_\Delta(X) = \bigcup Equiv_{\Delta-1}(X_i).
$
We can associate every \textDelta-class of equivalence with the size of its passkeys. The corresponding size is called \textit{level} of the class of equivalence. The smaller the level is, the simpler is the class of equivalence. Then for any \textDelta{} the complexity of the dataset can be captured by means of the distribution of the classes of equivalence within the levels, called here \textit{a level structure}.




%


Let us associate the following measures to an itemset $X$:
\begin{align}
    \Delta(X)  &=\max\{0 \leq \Delta \leq |G| \mid X \text{ is \textDelta-closed}\}\label{eq:delta} \\
    \Delta_{key}(X) &= \max\{0 \leq \Delta \leq |G| \mid X \text{ is a \textDelta-key}\} \\
    \Delta_{pk}(X) &= \max\{0 \leq \Delta \leq |G| \mid X \text{ is a \textDelta-passkey}\}
\end{align}

All these measures capture the maximal \textDelta{} for which the itemset $X$ preserves a certain property, i.e.,  being closed, being a key, or a passkey. We will call these values $\Delta$-values of closed itemsets, keys, and passkeys, respectively.
Since any passkey is a key, $\Delta_{passkey}(X) \leq \Delta_{key}(X)$.
It is clear from definitions that for all $1 \leq \Delta \leq \Delta_{key}(X)$, $X$ is a \textDelta-key, and for all $1 \leq \Delta \leq \Delta_{passkey}(X)$, $X$ is a \textDelta-passkey.
%
Let us now relate this measures to variations of a dataset. \textit{Would it be possible to higihlight the itemsets that are likely to be preserved if a dataset is changed?} 

\begin{proposition}\label{prop:passkey-under-sample}
    Let $\Delta_{passkey}(X)=\delta$, then at least $\delta$ objects should be removed from the dataset in order to have a subdataset $\K_s$ such that $X$ is not a passkey in $\K_s$.
\end{proposition}

\begin{proposition}\label{prop:key-under-sample}
    Let $\Delta_{key}(X)=\delta$, then at least $\delta$ objects should be removed from the dataset in order to obtain a subdataset $\K_s$ such that $X$ is not a key in $\K_s$.
\end{proposition}
\begin{proposition}\label{prop:closure-under-sample}
    Let $\Delta(X)=\delta$, then at least $\delta$ objects should be removed from the dataset in order to obtain a subdataset $\K_s$ such that $X$ is not closed in $\K_s$.
\end{proposition}




The proofs can be found in the technical report\footnote{\url{https://arxiv.org/TO-BE-DONE}}.
The propositions shows that, the higher is \textDelta, the deeper should the dataset be modified for removing a certain property of the dataset.
Thus, if the structure of the dataset is captured with itemsets of high \textDelta, then the discovered itemsets are stable w.r.t. object removal and this is more related to the distribution underlying the dataset the objects are taken from.

Let us notice that an itemset $X$ is a \textDelta-key only if $\forall Y \subset X \left( |Y'| - |X'| \geq \Delta\right)$.
%
%
It is only a necessary condition.
Indeed, consider the following context with 3 objects $G=\{g_1,g_2,g_3\}$, $\{g_1\}'=\emptyset$, $\{g_2\}'=\{m_1\}$, and $\{g_3\}'=\{m_1,m_2\}$ then the aforementioned condition 
is satisfied for $\{m_2\}$ and $\Delta=2$.
However, $\{m_2\}$ is not a \textDelta-key.
This condition 
corresponds to \textdelta-free sets~\cite{BoulicautBR03}. Thus, not every \textdelta-free set is a \textDelta-key.

\subsection{Example}\label{sect:example}

\begin{figure}
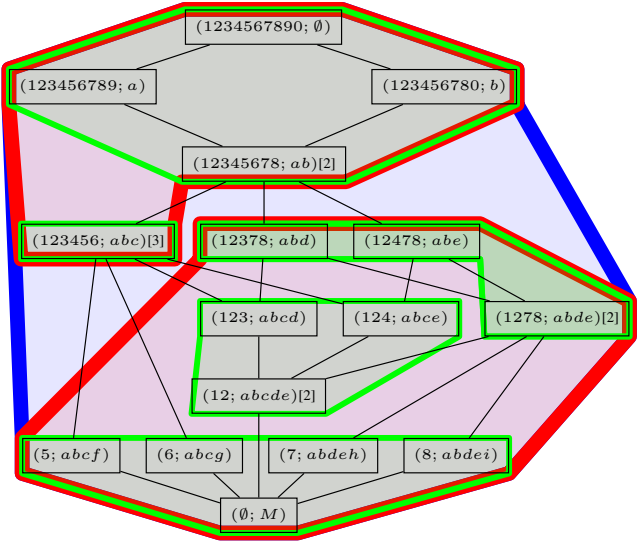

    \centering
    \resizebox{\columnwidth}{!}{
        \putDeltaLattice
    }
    \caption{
        The hierarchical structure of \textDelta-equivalence classes.
    }
    \label{fig:equiv-for-deltas}
\end{figure}

In Fig.~\ref{fig:stability-context} a formal context is shown.
Every object except for object 1 has its own attribute and thus the size of their passkey is 1. The closed itemsets (intents) along with their supports (extents) are shown in Fig.~\ref{fig:equiv-for-deltas}.
For brevity sake, we write 123 instead of $\{1,2,3\}$ and so on. The concepts with extents 123456 and 12478 have passkeys equal to $c$ and $e$, respectively.
The passkey for concept with extent 12345678 is $ab$ and the passkey for the top concept is $\emptyset$.
The concept with a passkey larger than 2 is the concept with extent 12.


In Fig.~\ref{fig:equiv-for-deltas} in the square brackets, the \textDelta{} of the intent, called hereafter \textDelta-measure, indicated for every concept if it is different from 1.
The concept intents are the maximal elements in the \textDelta-classes of equivalence for \textDelta{} not larger then their \textDelta-measure.

Let us find the \textDelta-classes of equivalence for $\Delta=2$.
There are only 4 concepts with \textDelta-measure not smaller than 2, thus, there are only 4 classes of equivalence for $\Delta=2$ plus the technical class of equivalence at the bottom.
These new classes of equivalence are formed by joining smaller classes of equivalence, i.e., the 1-classes of equivalence related to single concepts.
In Fig.~\ref{fig:equiv-for-deltas} the classes of equivalence for $\Delta=2$ are shown by green areas, while the concepts in the figure are 1-classes of equivalence.

Now, if we consider \textDelta-classes of equivalence with $\Delta=3$, there are only two closures: $abc$ and $\A$.
The class where the maximal itemset is $abc$ contains the concepts with the following intents: $abc$, $ab$, $a$, $b$, $\emptyset$.
The other concepts belong to the class of equivalence whose maximal element is $\A$.
If we further increase \textDelta, the whole lattice will collapse into a single class of equivalence. 
In Fig.~\ref{fig:equiv-for-deltas} these classes of equivalence are shown with red areas.
%
\def\drawpolygon#1,#2,#3,#4;{
    \begin{pgfonlayer}{background}
        \filldraw[line width=#2,join=round,#1,fill opacity = 0.1](#3)foreach\A in{#4}{--(\A)}--cycle;
    \end{pgfonlayer}
}

\tikznamedpicture{\putDeltaLattice}{[
	every node/.style={draw,rectangle,font={\tiny}},
	node distance= 0.5cm and 0.3cm
	]
	\node(g5){\concept{5}{abcf}};
	\node(g6)[right=of g5]{\concept{6}{abcg}};
	\node(g7)[right=of g6]{\concept{7}{abdeh}};
	\node(g8)[right=of g7]{\concept{8}{abdei}};
	\node(g12)[above=of $(g6)!0.5!(g7)$] {\concept{12}{abcde}[2]};
	\node(bottom)[below=of $(g6)!0.5!(g7)$] {\concept{\emptyset}{\A}}
	edge(g5)
	edge(g6)
	edge(g7)
	edge(g8)
	edge(g12);
	\node(g123)[above=of g12] {\concept{123}{abcd}}
	edge(g12);
	\node(g124)[right=of g123] {\concept{124}{abce}}
	edge(g12);
	\node(g1278)[right=of g124] {\concept{1278}{abde}[2]}
	edge(g7)
	edge(g8)
	edge(g12);
	\node(g123456)[above left=of g123] {\concept{123456}{abc}[3]}
	edge(g5)
	edge(g6)
	edge(g123)
	edge(g124);
	\node(g12378)[right=of g123456] {\concept{12378}{abd}}
	edge(g123)
	edge(g1278);
	\node(g12478)[right=of g12378] {\concept{12478}{abe}}
	edge(g124)
	edge(g1278);
	\node(g12345678)[above=of g12378] {\concept{12345678}{ab}[2]}
	edge(g123456)
	edge(g12378)
	edge(g12478);
	\node(g123456789)[above left=of g12345678] {\concept{123456789}{a}}
	edge(g12345678);
	\node(g123456780)[above right=of g12345678] {\concept{123456780}{b}}
	edge(g12345678);
	\node(top)[above=of $(g123456789)!0.5!(g123456780)$] {\concept{1234567890}{\emptyset}}
	edge(g123456789)
	edge(g123456780);
	
	\drawpolygon blue,5, top.north west, top.north east, g123456780.north east, g123456780.south east, g1278.north east, g1278.south east, g8.south east, bottom.south east, bottom.south west, g5.south west, g5.north west, g123456789.south west, g123456789.north west;
	
	g12345678.south east, g12345678.south west,g123456.north east,g123456.south east,g123456.south west, g123456789.south west, g123456789.north west;
	
	\drawpolygon red,5, top.north west, top.north east, g123456780.north east, g123456780.south east, g12345678.south east, g12345678.south west,g123456.north east,g123456.south east,g123456.south west, g123456789.south west, g123456789.north west;
	\drawpolygon red,5, g5.north west,g5.south west,bottom.south west,bottom.south east,g8.south east, g1278.south east, g1278.north east,g12478.north east,g12378.north west, g12378.south west;
	
	\drawpolygon green,2,g12378.north west,g12478.north east, g1278.north east, g1278.south east, g1278.south west, g12478.south east, g12378.south west;

	\drawpolygon green,2,g5.north west,g5.south west,bottom.south west,bottom.south east,g8.south east,g8.north east;
	\drawpolygon green,2,g12.south west,g12.south east,g124.south east, g124.north east, g123.north west, g123.south west;
	\drawpolygon green,2,g12378.north west,g12478.north east, g1278.north east, g1278.south east, g1278.south west, g12478.south east, g12378.south west;
	\drawpolygon green,2, g123456.north west,g123456.north east,g123456.south east,g123456.south west;
	\drawpolygon green,2, top.north west, top.north east, g123456780.north east, g123456780.south east, g12345678.south east, g12345678.south west, g123456789.south west, g123456789.north west;
}

\subsection{Computational considerations}
%

The algorithm for computing \textDelta-equivalence classes is sketched in Algorithm~\ref{alg:general} while
details are provided in Algorithm~\ref{alg:passkey-delta}.
First,  we compute the concepts and the associated concept levels using GDPM algorithm~\cite{MakhalovaBKN22}, which iterates over keys searching for passkeys.
Then, for every concept we compute the \textDelta-value of the concept intent given by equation~(\ref{eq:delta}).
Finally, for every concept we compute the \textDelta{} of the concept passkey.
This procedure is made precise in Algorithm~\ref{alg:passkey-delta}.




\begin{algorithm}[h]
    \SetAlgoLined
    \SetKwFunction{GDPM}{GDPM}
    \SetKwFunction{CDeltas}{ComputeDeltas}
    \SetKwFunction{CPKDeltas}{ComputePKDeltas}
    \KwData{$\K$ is the dataset}
    \KwResult{A set of concepts associated with their levels and the values of \textDelta{} for the concept intent and for the concept passkey}
    $\{c_i \Rightarrow level_i\} = \GDPM{$\K$}$\;
    $\{c_i \Rightarrow \left<\Delta_{cls}(c_i), Ref(c_i)\right>\} = \CDeltas{$\{c_i\}$}$\;
    $\{c_i \Rightarrow \Delta_{passkey}(c_i)\} = \CPKDeltas{$\{c_i \Rightarrow \left<level_i, \Delta_{cls}(c_i), Ref(c_i)\right>\}$}$\;
\caption{A general algorithm for finding \textDelta{}-values for all concept intents and for their passkeys.\label{alg:general}}
\end{algorithm}

\begin{algorithm}[h]
    \SetAlgoLined
    \KwData{$\{c_i\}$ is a set of concepts with associated levels, \textDelta{}-value for their intent and the most closest child}
    \KwResult{The value of \textDelta{} for passkeys of every concept}
    \SetKwProg{Def}{def}{:}{}
    \Def{\CPKDeltas{$\{c_i\}$}}{
        \ForEach{$c \in \{c_i\}$}{
            $\Delta_{pk}(c) = 1$\;
            $Cls(c) = c$
        }
        $d=1$\;
        \While{$d < |G|$}{
            $d = d+1$\;
            $hasUpdates = True$\;
            \While{$hasUpdates\text{ is }True$}{
                $hasUpdates = False$\;
                \ForEach{$c \in \{c_i\}$}{
                    $cls = Cls(c)$\;
                    \If{$|Ext(Ref(cls))| - |Ext(cls)| < d$}{
                       $Cls(c) = Cls(cls)$\;
                       $hasUpdates = True$\;
                    }
                }
            }
            \ForEach{$c \in \{c_i\}$}{
                $cls = Cls(c)$\;
                $level_\Delta(cls) = \min(level_\Delta(cls), level(c))$\;
            }
            \ForEach{$c \in \{c_i\}$}{
                $cls = Cls(c)$\;
                \If{$level(c) == level_\Delta(cls)$}{
                    $\Delta_{pk}(c) = d$\;
                }
            }
        }
    }

\caption{An algorithm for computing \textDelta{} of passkeys for every concept.\label{alg:passkey-delta}}
\end{algorithm}




The computational complexity of the algorithm shown in Algorithm~\ref{alg:passkey-delta} is $O(|G|\cdot \log(|G|) \cdot |\{c_i\}|)$ since we need to iterate over all possible thresholds of \textDelta{} and for any given threshold we need to find the \textDelta-closure of every concept that can be found for at most $\log(|G|)$ since on every new iteration of cycle 12--17 the distance to the closure from the concept is either duplicate or the closure is found. 


%
\section{Experiments}\label{sec:experiments}
%
%
The experiments are carried out on the system with Intel Core i5 CPU, 16GB of RAM and Nvidea GeForce RTX 3080 video card operated under Ubuntu 20.04 operating system. Table~\ref{tbl:computation-time} shows datasets with total computation time larger than 10 seconds. It also shows the number of objects, attributes, and closed itemsets. The computation time is divided into the  computation time of the level structure with GDPM and the time for computing \textDelta-classes of equivalence. We can see that the most important determinant of the computation time is the number of closed itemsets in the dataset, the second determinant being the size of the dataset.

\begin{table}[t]
    \centering
    \caption{Dataset characteristics and computation time}
    \label{tbl:computation-time}
    \begin{tabular}{c|ccc|cc}
        \hline\hline
         \multirow{2}{*}{Datasets} & \multirow{2}{*}{$|G|$} & \multirow{2}{*}{$|M|$} & \multirow{2}{*}{\# closed} & \multicolumn{2}{|c}{runtime, sec}   \\
         &&&& GDPM & \textDelta-passkeys \\
         \hline
        adult & 48842 & 95 & 359141 & 984 & 247 \\
        chess kr k & 28056 & 40 & 84636 & 146 & 52 \\
        cyl. bands &  540 & 120 & 39829537 & 2404 & 8803 \\
        horse col. &  368 & 81 & 173866 & 11 & 21 \\
        ionosphere & 351 & 155 & 23202541 & 2467 & 3090 \\
        mushroom  & 8124 & 88 & 181945 & 164 & 27 \\
        nursery & 12960 & 27 & 115200 & 46 & 24 \\
        pen digits & 10992 & 76 & 3605507 & 12863 & 5020 \\
        soybean & 683 & 99 & 2874252 & 379 & 393 \\
        \hline\hline
    \end{tabular}
\end{table}

\subsection{Stability of passkeys.}
The goal of the experiments is to verify that \textDelta-closure structures are becoming more robust when \textDelta\, is increasing.
In particular, \textDelta-closure structures when \textDelta\, is greater than 1 are more robust than closure structures or 1-closure structures.

For any \textDelta-equivalence class we distinguish its minimum elements called the \textDelta-passkeys and the maximum one called the \textDelta-closed itemset.
They are both characterized by \textDelta, where the higher value for \textDelta, the more stable are passkeys/closed itemsets w.r.t. noise.
A passkey can be considered as a most concise ``definition'' of a \textDelta-closed itemset.

In this section we study how stable \textDelta-equivalence classes are distributed in the dataset. 
To analyze the datasets we use the following observations:
\begin{itemize}
    \item[(i)] 
 the equivalence classes with small passkeys are easier to mine by the algorithm that computes the closure structure~\citep{MakhalovaBKN22}, and easier to interpret, since \textDelta-passkeys are the smallest elements in the \textDelta-equivalence classes,
    \item[(ii)]
 the \textDelta-equivalence classes where passkeys have high \textDelta\, are more stable w.r.t. noise.
\end{itemize}

\begin{figure*}
    \begin{minipage}[c]{.26\textwidth}
        \includegraphics[width=\textwidth]{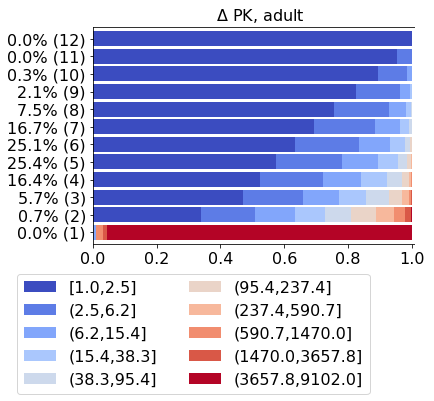}
    \end{minipage}\hspace{-2em}
    \begin{minipage}[c]{.26\textwidth}
        \includegraphics[width=\textwidth]{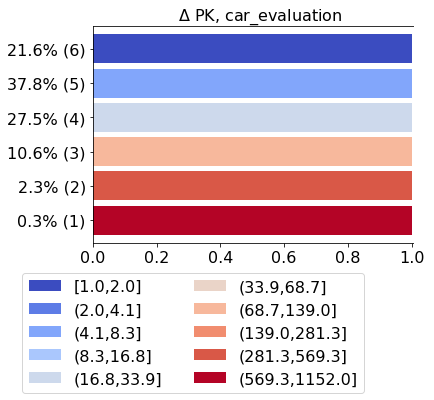}
    \end{minipage}\hspace{-2em}
    \begin{minipage}[c]{.26\textwidth}
        \includegraphics[width=\textwidth]{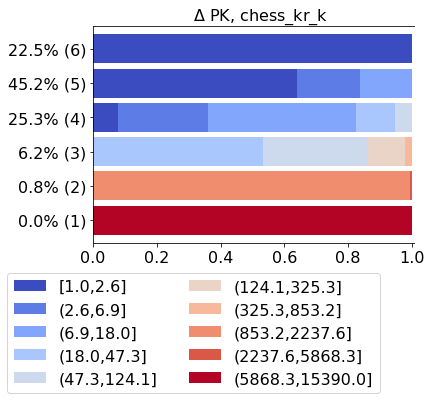}
    \end{minipage}\hspace{-2em}
    \begin{minipage}[c]{.26\textwidth}
        \includegraphics[width=\textwidth]{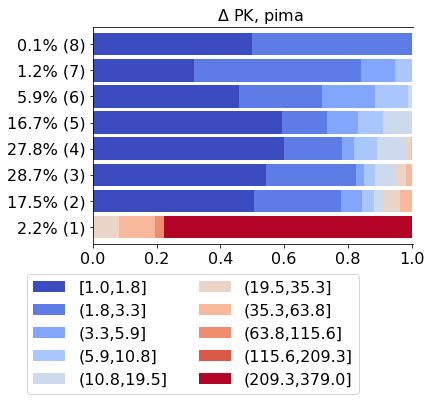}
    \end{minipage}
    \begin{minipage}[c]{.24\textwidth}
        \centering
        (a) typical
    \end{minipage}
    \begin{minipage}[c]{.24\textwidth}
        \centering
        (b) deterministic
    \end{minipage}
    \begin{minipage}[c]{.24\textwidth}
        \centering
        (c) intermediate
    \end{minipage}
    \begin{minipage}[c]{.24\textwidth}
        \centering
        (d) intermediate
    \end{minipage}
    \caption{%
    The distribution of \textDelta-passkeys by levels based on 10 frequency bins distributed within the \textDelta-closure levels. 
    The horizontal bars represent closure levels, where the level number is given in parentheses.
    The leftmost value is the percentage of closed itemsets $|\mathcal{C}_{k}|/|\mathcal{C}|$ at level $k$.
    The width of a color within a bar is proportional to the ratio of \textDelta-passkeys for \textDelta{} in $(v_1, v_2]$ at level $\mathcal{C}_k$.
    }
    \label{fig:delta}
\end{figure*}

We analyzed 25 datasets from LUCS-KDD repository (their parameters are given in the supplementary material) paying attention to the properties mentioned above.
Among all the datasets under analysis we discovered 4 main types of behaviors which are displayed in the columns of~Fig.~\ref{fig:delta}.

\subsection{Visualizing the closure structure.}

Fig.~\ref{fig:delta} visualizes \textDelta-closure-structures for different types of data.
%
The first level at the bottom of each subfigure shows the distribution of passkeys constituted of only one attribute w.r.t. their values of \textDelta.
Actually this first level corresponds to the attribute frequency distribution and does not contain any information about interaction between attributes.
By contrast, starting from the second level, the \textDelta-closure levels include information about ``interactions" between attributes.
For example, the second level contains \textDelta-closed itemsets generated from passkeys of size 2.

Datasets may have mainly 3 different types of behaviors.
A majority of datasets among those that were analyzed, namely ``auto'', ``cylinder bands'', ``dermatology'', ``ecoli'', ``glass'', ``heart diseases'', ``hepatitis'', ``horse colic'', ``pen digit'', ``soybean'', ``wine'', ``zoo'', have a closure structure which is similar to the closure structure of ``adult'', as displayed in Fig.~\ref{fig:delta}a.
We call this  type of behavior ``typical''.
For all these datasets, the ratio of the equivalence classes having passkeys with a low \textDelta\, increases exponentially with the level number.
It means that most of closed itemsets are not stable and can easily become not closed if new objects are added.
The equivalence classes which are the most stable to noise are located at the first levels.

The second type of behavior is given in Fig.~\ref{fig:delta}b and termed as ``deterministic''.
It is observed for ``car evaluation'' and ``nursery'' datasets, where each level mostly contains \textDelta-equivalence classes with passkeys having the same value. 
Actually these datasets were built from hierarchical decision models.
Then each entry in such a dataset usually has a rather ``determinististic'' behavior, i.e., a much less random nature while higher order interactions between attributes are unusual.

The ``intermediate'' behavior can be different. In some cases  the number of ``unstable'' \textDelta-equivalence classes grows less rapidly than for the majority of ``typical'' datasets.
This means that a larger number of ``stable'' \textDelta-equivalence classes are present in the upper levels.
The other datasets with similar behavior are ``iris'', ``led7'', ``mushroom'', and ``tic tac toe''.

Another type of intermediate behavior was observed for the datasets ``ionosphere'', ``page blocks'', and ``pima'', and is presented in Fig.~\ref{fig:delta}d, where the \textDelta-closure structure of ``pima'' is shown.
For these datasets, the ratio of the \textDelta-equivalence classes including passkeys with larger \textDelta-values may increase with levels, i.e., as the levels increase, the ratio of more robust \textDelta-equivalence classes increases as well.
Intuitively, this kind of behavior is related to datasets where the 1st level of the \textDelta-closure structure contains a substantial number of attributes of very low frequency.
This induces unreliable or unstable itemsets in the upper levels of the \textDelta-closure structure.



Finally, the  patterns with high \textDelta-values  are even more interesting if they are located at high levels of the closure structure, since they reflect high order interactions between attributes and are very noise-stable.

\section{Discussion and Conclusion}
\label{sec:conclusion}

In this paper we introduced the ``closure structure'' and the ``\textDelta-closure structure'' of a dataset, along with their theoretical and practical properties.
The closure structure provides a view of the content of a dataset and may be used for guiding data mining.
We have found three types of ``dataset behaviors'' which are showing the stability and robustness of the corresponding closed itemsets.
Here after we synthesize the main aspects of the closure structures:

\begin{itemize}
\item
  The 
  \textDelta-closure structure is determined by the 
  \textDelta-closed itemsets and their equivalence classes, which
   make a partition of the whole set of itemsets and are represented by \textDelta-passkeys, i.e., the smallest keys in an equivalence class. Accordingly, passkeys can represent any closed itemset of attributes without loss.
\item
  The closure structure is organized around a set of levels depending on the size of the passkeys, which is related to the distribution of the closed itemsets and their passkeys.
  Moreover, the \textDelta-closure structure allows us to study the stability and robustness of the closed itemsets.
%
\end{itemize}

  We should notice that more studies should be done for deeper understanding how the strategy based on the \textDelta-closure structure can help us to deal with noisy data, and how to practically take into account the noise in the analysis of data and the interpretation of the resulting patterns.

\section*{Acknowledgments}
Sergei O. Kuznetsov  would like to thank Basic Research Program of the HSE University for its support.
\bibliographystyle{siam}
\bibliography{icdm22}

\begin{thebibliography}{10}

\bibitem{AggarwalH14}
{\sc C.~C. Aggarwal and J.~Han}, eds., {\em {Frequent Pattern Mining}},
  Springer, 2014.

\bibitem{BoleyHW09}
{\sc M.~Boley, T.~Horv{\'{a}}th, and S.~Wrobel}, {\em {Efficient Discovery of
  Interesting Patterns Based on Strong Closedness}}, in {Proceedings of the
  SIAM International Conference on Data Mining, (SDM)}, {SIAM}, 2009,
  pp.~1002--1013.

\bibitem{BoulicautBR03}
{\sc J.-F. Boulicaut, A.~Bykowski, and C.~Rigotti}, {\em {Free-sets: a
  condensed representation of boolean data for the approximation of frequency
  queries}}, {Data Mining and Knowledge discovery}, 7 (2003), pp.~5--22.

\bibitem{ChengKN06}
{\sc J.~Cheng, Y.~Ke, and W.~Ng}, {\em {$\delta$-Tolerance Closed Frequent
  Itemsets}}, in {Proceedings of the 6th IEEE International Conference on Data
  Mining (ICDM)}, {IEEE} Computer Society, 2006, pp.~139--148.

\bibitem{DzyubaLR17}
{\sc V.~Dzyuba, M.~van Leeuwen, and L.~D. Raedt}, {\em {Flexible constrained
  sampling with guarantees for pattern mining}}, {Data Mining and Knowledge
  Discovery}, 31 (2017), pp.~1266--1293.

\bibitem{FlouvatMP10}
{\sc F.~Flouvat, F.~D. Marchi, and J.~Petit}, {\em {A new classification of
  datasets for frequent itemsets}}, Journal of Intelligent Information Systems,
  34 (2010), pp.~1--19.

\bibitem{GanterW99}
{\sc B.~Ganter and R.~Wille}, {\em {Formal Concept Analysis -- Mathematical
  Foundations}}, Springer, 1999.

\bibitem{KlimushkinOR10}
{\sc M.~Klimushkin, S.~A. Obiedkov, and C.~Roth}, {\em {Approaches to the
  Selection of Relevant Concepts in the Case of Noisy Data}}, in {Proceedings
  of the 8th International Conference on Formal Concept Analysis (ICFCA)},
  L.~Kwuida and B.~Sertkaya, eds., Lecture Notes in Computer Science 5986,
  Springer, 2010, pp.~255--266.

\bibitem{MakhalovaBKN22}
{\sc T.~Makhalova, A.~V. Buzmakov, S.~O. Kuznetsov, and A.~Napoli}, {\em
  {Introducing the closure structure and the GDPM algorithm for mining and
  understanding a tabular dataset}}, International Journal of Approximate
  Reasoning, 145 (2022), pp.~75--90.

\bibitem{PasquierBTL99}
{\sc N.~Pasquier, Y.~Bastide, R.~Taouil, and L.~Lakhal}, {\em Efficient mining
  of association rules using closed itemset lattices}, Information Systems, 24
  (1999), pp.~25--46.

\bibitem{PensaB05}
{\sc R.~G. Pensa and J.~Boulicaut}, {\em {Towards Fault-Tolerant Formal Concept
  Analysis}}, in {Proceedings of the 9th Congress of the Italian Association
  for Artificial Intelligence (AI*IA)}, S.~Bandini and S.~Manzoni, eds.,
  Lecture Notes in Computer Science 3673, Springer, 2005, pp.~212--223.

\bibitem{RameshMZ03}
{\sc G.~Ramesh, W.~Maniatty, and M.~J. Zaki}, {\em {Feasible itemset
  distributions in data mining: theory and application}}, in {Proceedings of
  the 22nd ACM SIGACT-SIGMOD-SIGART Symposium on Principles of Database Systems
  (PODS)}, F.~Neven, C.~Beeri, and T.~Milo, eds., {ACM}, 2003, pp.~284--295.

\bibitem{RiondatoV14}
{\sc M.~Riondato and F.~Vandin}, {\em {Finding the True Frequent Itemsets}}, in
  {Proceedings of the SIAM International Conference on Data Mining (SDM)},
  M.~J. Zaki, Z.~Obradovic, P.~Tan, A.~Banerjee, C.~Kamath, and
  S.~Parthasarathy, eds., {SIAM}, 2014, pp.~497--505.

\bibitem{SmetsV12}
{\sc K.~Smets and J.~Vreeken}, {\em {Slim: Directly Mining Descriptive
  Patterns}}, in {Proceedings of the Twelfth SIAM International Conference on
  Data Mining (SDM)}, {SIAM} / Omnipress, 2012, pp.~236--247.

\bibitem{LeeuwenU14}
{\sc M.~van Leeuwen and A.~Ukkonen}, {\em {Fast Estimation of the Pattern
  Frequency Spectrum}}, in {Proceedings of the European Conference on Machine
  Learning and Knowledge Discovery in Databases (ECML-PKDD)}, T.~Calders,
  F.~Esposito, E.~H{\"{u}}llermeier, and R.~Meo, eds., Lecture Notes in
  Computer Science 8725, Springer, 2014, pp.~114--129.

\bibitem{VreekenT14}
{\sc J.~Vreeken and N.~Tatti}, {\em {Interesting Patterns}}, in {Frequent
  Pattern Mining}, C.~C. Aggarwal and J.~Han, eds., Springer, 2014,
  pp.~105--134.

\bibitem{YangFB01}
{\sc C.~Yang, U.~M. Fayyad, and P.~S. Bradley}, {\em {Efficient discovery of
  error-tolerant frequent itemsets in high dimensions}}, in {Proceedings of the
  seventh ACM SIGKDD International Conference on Knowledge Discovery and Data
  Mining (KDD)}, D.~Lee, M.~Schkolnick, F.~J. Provost, and R.~Srikant, eds.,
  {ACM}, 2001, pp.~194--203.

\end{thebibliography}

\end{document}


\section*{8. Supplementary material}

\begin{table}[h]
    \caption{Parameters of the analyzed datasets}
    \centering
    \setlength{\tabcolsep}{.5pt}
    \begin{tabular}{l|ccccc}
    \toprule
    name & $|G|$ & $|M|$ & \#attr. & \begin{tabular}[c]{@{}c@{}}closure\\index\\$CI$\end{tabular}& \begin{tabular}[c]{@{}c@{}}\#closed\\concepts\end{tabular} \\\midrule
    adult & 48842 & 95 & 14 & 12 & 359141  \\
    auto & 205 & 129 & \underline{\textbf{8}}&\underline{\textbf{8}}& 57789 & 0 & 1 & 3 & 6& 3 & \textbf{3}\\
    breast & 699 & 14 & 10 & 6 & 361\\
    car eval. & 1728 & 21 & \underline{\textbf{6}} & \underline{\textbf{6}} & 8000 \\
    chess kr k & 28056 & 40 & \underline{\textbf{6}} & \underline{\textbf{6}} & 84636\\
    cyl. bands & 540 & 120 & 39 & 19 &39829537\\
    dermat. & 366 & 43 & 33 & 9 & 14152\\
    ecoli & 327 & 24 & 8 & 6 & 425\\
    glass & 214 & 40 & 10 & 8 & 3246\\
    heart-dis. & 303 & 45 &  & 9 & 25539\\
    hepatitis & 155 & 50 & 19 & 11 & 144871\\
    horse colic & 368 & 81 & 27 & 9 & 173866\\
    ionosphere & 351 & 155 & 34 & 17 & 23202541\\
    iris & 150 & 16 & \underline{\textbf{4}} & \underline{\textbf{4}} & 120\\
    led7 & 3200 & 14 & \underline{\textbf{7}} & \underline{\textbf{7}} & 1951\\
    mushroom & 8124 & 88 & 22 & 10 & 181945\\
    nursery & 12960 & 27 & \underline{\textbf{8}} & \underline{\textbf{8}} & 115200\\
    page blocks & 5473 & 39 & 10 & 8 & 723\\
    pen digits & 10992 & 76 & 16 & 11 & 3605507\\
    pima & 768 & 36 & \underline{\textbf{8}} & \underline{\textbf{8}} & 1626\\
    soybean & 683 & 99 & 35 & 13 & 2874252\\
    tic tac toe & 958 & 27 & 9 & 7 & 42712\\
    wine & 178 & 65 & 13 & 7 & 13229\\
    zoo & 101 & 35 & 17 & 7 & 4570
    \\\bottomrule
    \end{tabular}
\end{table}

\begin{figure}
     \begin{minipage}[c]{.3\textwidth}
        \includegraphics[width=\textwidth]{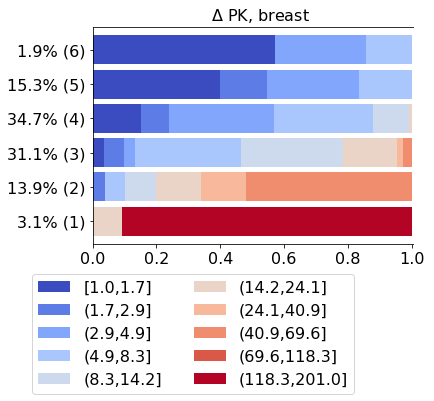}
     \end{minipage}
     \begin{minipage}[c]{.3\textwidth}
        \includegraphics[width=\textwidth]{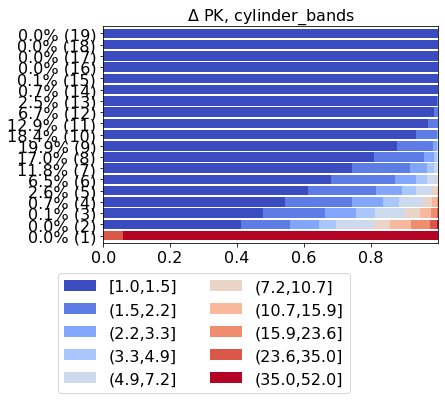}
     \end{minipage}
     \begin{minipage}[c]{.3\textwidth}
        \includegraphics[width=\textwidth]{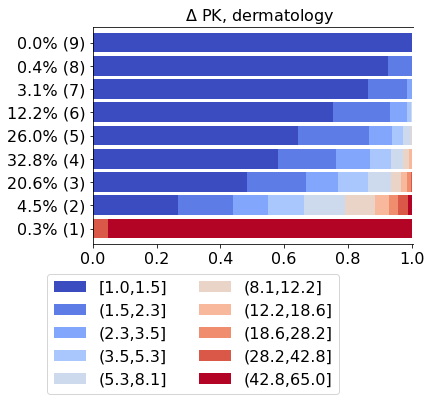}
     \end{minipage} 
     \begin{minipage}[c]{.3\textwidth}
        \includegraphics[width=\textwidth]{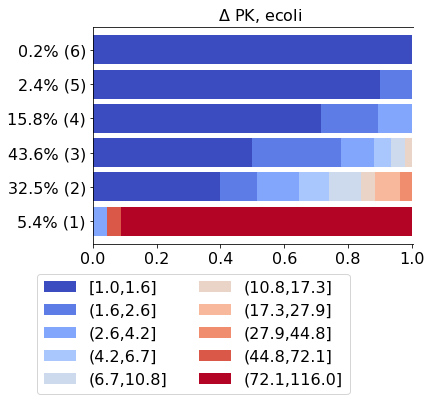}
     \end{minipage} 
     \begin{minipage}[c]{.3\textwidth}
        \includegraphics[width=\textwidth]{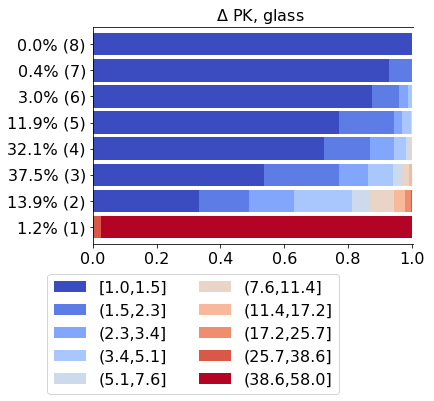}
     \end{minipage} 
     \begin{minipage}[c]{.3\textwidth}
        \includegraphics[width=\textwidth]{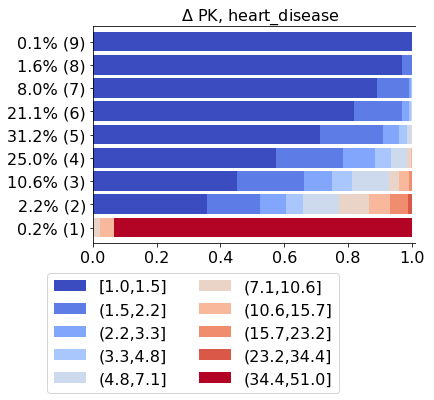}
     \end{minipage} 
     \begin{minipage}[c]{.3\textwidth}
        \includegraphics[width=\textwidth]{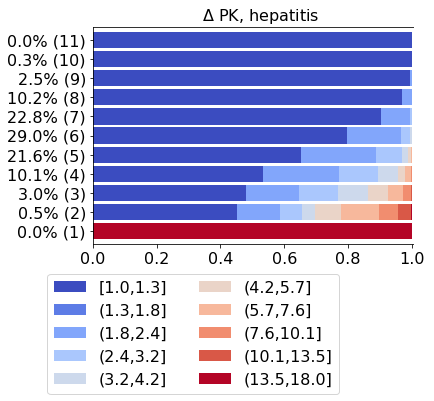}
     \end{minipage} 
     \begin{minipage}[c]{.3\textwidth}
        \includegraphics[width=\textwidth]{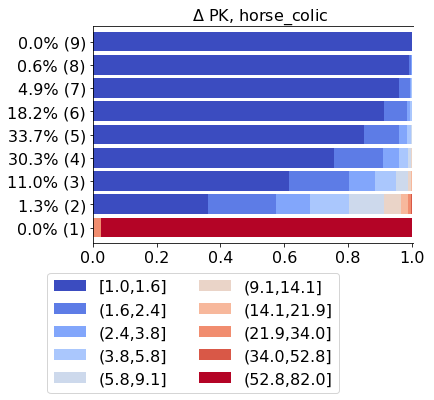}
     \end{minipage} 
     \begin{minipage}[c]{.3\textwidth}
        \includegraphics[width=\textwidth]{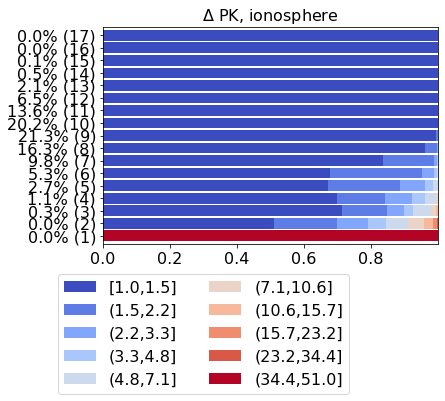}
     \end{minipage} 
     \begin{minipage}[c]{.3\textwidth}
        \includegraphics[width=\textwidth]{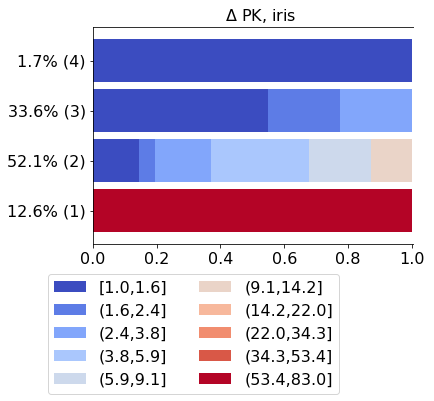}
     \end{minipage} 
     \begin{minipage}[c]{.3\textwidth}
        \includegraphics[width=\textwidth]{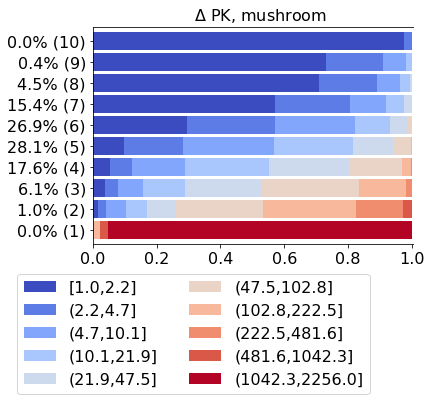}
     \end{minipage} 
     \begin{minipage}[c]{.3\textwidth}
        \includegraphics[width=\textwidth]{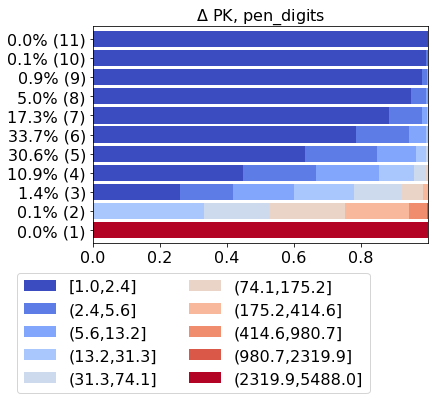}
     \end{minipage} 
     \begin{minipage}[c]{.3\textwidth}
        \includegraphics[width=\textwidth]{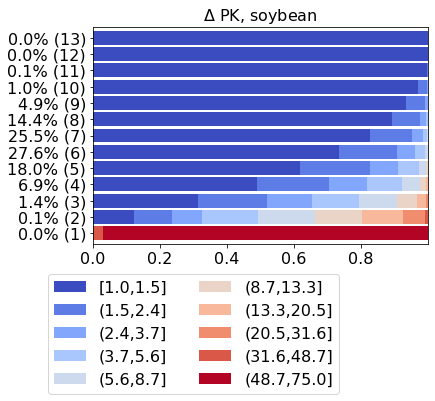}
     \end{minipage} 
     \begin{minipage}[c]{.3\textwidth}
        \includegraphics[width=\textwidth]{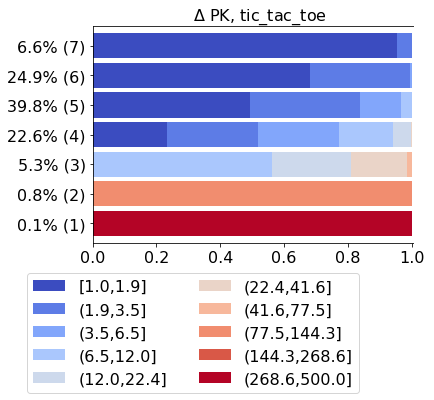}
     \end{minipage} 
     \begin{minipage}[c]{.3\textwidth}
        \includegraphics[width=\textwidth]{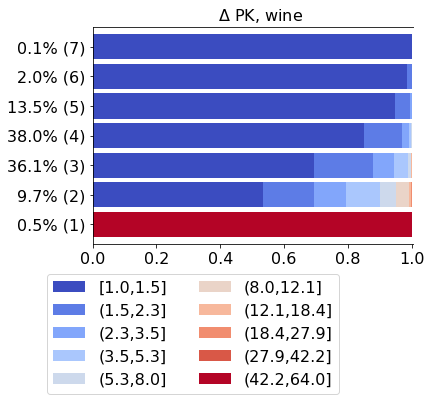}
     \end{minipage} 
     \begin{minipage}[c]{.3\textwidth}
        \includegraphics[width=\textwidth]{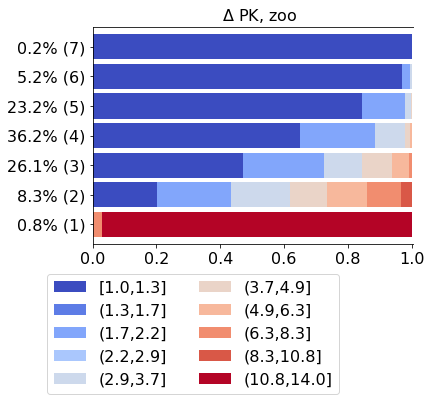}
     \end{minipage} 
    \caption{The $\Delta$-equivalence classes for the analyzed datasets}
\end{figure}